\documentclass[conference,letterpaper]{IEEEtran}
\usepackage{cite}
\usepackage{amsmath,amssymb}
\usepackage[hidelinks]{hyperref}
\usepackage{microtype}
\usepackage{graphicx} 

\title{Learning Which Device to Call: Online Tool Grounding for LLM Agents in IoT}

\author{
\IEEEauthorblockN{\normalsize Yichen Liu\textsuperscript{1}, Xiaochong Jiang\textsuperscript{2}, Haoran Yu\textsuperscript{1}, Lifei Liu\textsuperscript{1}}
\IEEEauthorblockA{\small \textsuperscript{1}Independent Researcher, Seattle, WA, USA\quad \textsuperscript{2}Independent Researcher, Boston, MA, USA\\
Email: yil160@ucsd.edu, jiang.xiaoc@northeastern.edu, haoranyu889@gmail.com, lliu.lifei@gmail.com}
}

\begin{document}
\maketitle

\begin{abstract}
The Internet of Things (IoT) gives large language model (LLM) agents access to
physical observations, but one requested capability is usually offered by
several devices, and the agent's plan does not say which one to call. Existing
work retrieves relevant tools and checks plan feasibility, without learning
which eligible device returns a useful observation. Unlike previous works, we
study online tool grounding: binding an already specified tool call to one
physical endpoint under service cost and device participation constraints. We
propose a framework that learns which physical device should execute each
agent-issued call. It formulates selection as a contextual bandit problem and
combines adaptive context partitions with hierarchical endpoint trees, so that
execution feedback is reused across related requests and devices. Selection
balances task success against service cost while respecting device capability,
authorization, battery reserves, and availability. Gaussian reward aggregation
provides an optional mechanism for protecting individual feedback values.
Theoretical results establish a finite-horizon regret bound and differential
privacy for released reward statistics under fixed, reward-independent
assignments. In a simulated loading-dock workload, the proposed router achieves
higher step success and better cost-adjusted service, at lower simulated device
energy and latency, than nearest-device, advertised-quality, UCB1, and LinUCB
selection.
\end{abstract}


\section{Introduction}
Large language model (LLM) agents provide a way to interact with Internet of Things (IoT) systems through natural-language requests. By combining reasoning with tool calls, an agent can gather observations, assess the returned evidence, and determine its next action~\cite{yao2023react}. The agent, however, reasons about capabilities, whereas the deployment consists of individual devices.

Consider an agent asked whether a loading dock is blocked. The agent can issue \texttt{capture\_image(area)}, but this call does not identify a unique camera. A fixed camera and a robot-mounted camera may both cover the area, yet differ in viewing angle and image quality, so the system must choose which device to invoke. We call the assignment of an already specified tool call to one physical device endpoint \emph{device-bound tool grounding}, and we refer to each device--capability pair as an \emph{endpoint}.

Existing approaches address tool selection and physical feasibility from complementary directions: retrieval methods match requests to tool descriptions and adapt from interaction feedback~\cite{pan2025online}, while spatial and resource-aware planners verify that a sensing plan is executable~\cite{zhou2026iotbrain}. However, three problems remain. First, identifying a relevant capability or a feasible plan does not determine which eligible endpoint will deliver the most useful observation. A high-resolution camera may have an obstructed view, whereas a camera with lower resolution may provide sufficient evidence from a clearer angle~\cite{soro2009survey,soro2007camera}, and sensor outputs also degrade with device faults and operating conditions~\cite{goknil2023dataquality}. Usefulness is therefore a property of the request--endpoint pair and has to be learned from execution outcomes. Because only the selected device produces feedback, the router must balance using established endpoints against exploring alternatives.

Second, the choice must reflect the devices' current operating conditions. A mobile camera may move out of range, a battery-powered device may lack sufficient energy, and even a useful observation may arrive too late for the deadline. Selecting devices solely by past success therefore cannot ensure that the next call is feasible, and it depletes the reserves of the endpoints it favors.

Third, the execution history used to improve these decisions is itself sensitive. A sequence of feedback records describes when a particular device at a particular location was called and how it responded, which reveals activity patterns of the monitored space. Releasing individual rewards or unprotected aggregates, for example to an operator dashboard, therefore discloses more than routing requires, and the learning process needs a disclosure-limiting mechanism with a defined scope.

To address these challenges, we propose an online tool-grounding framework for LLM agents in connected IoT systems. We formulate each routing decision as a contextual bandit problem: the requested capability, target, quality requirements, and deadline define the context, and each eligible endpoint represents a candidate arm. The router checks compatibility, authorization, and predicted feasibility, selects one endpoint, and updates its utility estimate from a reward that combines task success with latency and energy cost. The LLM remains fixed: it chooses the operation and its target, and the router chooses the device.

The proposed learner uses adaptive context partitions and hierarchical endpoint trees to organize this experience. Context cells group requests with similar requirements, while the endpoint tree of each cell groups devices by their operating features, so that nearby request--endpoint pairs can share feedback under the stated smoothness assumptions. Refinement creates finer partitions as evidence accumulates. For privacy preservation, an optional Gaussian mechanism perturbs temporal-tree reward aggregates~\cite{dwork2014privacy}. The guarantee protects individual reward values in released statistics when metadata, record assignments, and release schedules are fixed independently of rewards, and it does not extend to the complete adaptive routing interaction.

The contribution of this paper lies in the device-bound tool-grounding setting and its formulation, not in new partition or aggregation components. The context and endpoint partitions follow the two-sided structure of~\cite{liu2022private}, and the private aggregation follows tree-based private learning and Gaussian summation~\cite{zhao2019privatetree,dwork2014privacy}. What is new is how these parts are combined for agent tool calls: eligibility is checked as a hard constraint before learning, the reward comes from verified tool-step outcomes and service cost, and the privacy statement names exactly which released statistics it covers.

\section{Related Work}
\label{sec:related-work}

\subsection{LLM Tool Use and Learning from Execution Feedback}
LLM agents interleave reasoning with tool use~\cite{yao2023react,yuan2026sp}, organize the evidence and memory that such workflows accumulate~\cite{zhu2026fundapod,chen2026survey,xu2025sedm}, and generate or retrieve candidate tool calls~\cite{patil2024gorilla,zheng2024toolrerank}. All of these decide \emph{which operation} to perform; our setting begins once the operation and target are fixed.

\begin{figure*}[t]
    \centering
    \includegraphics[width=\textwidth]{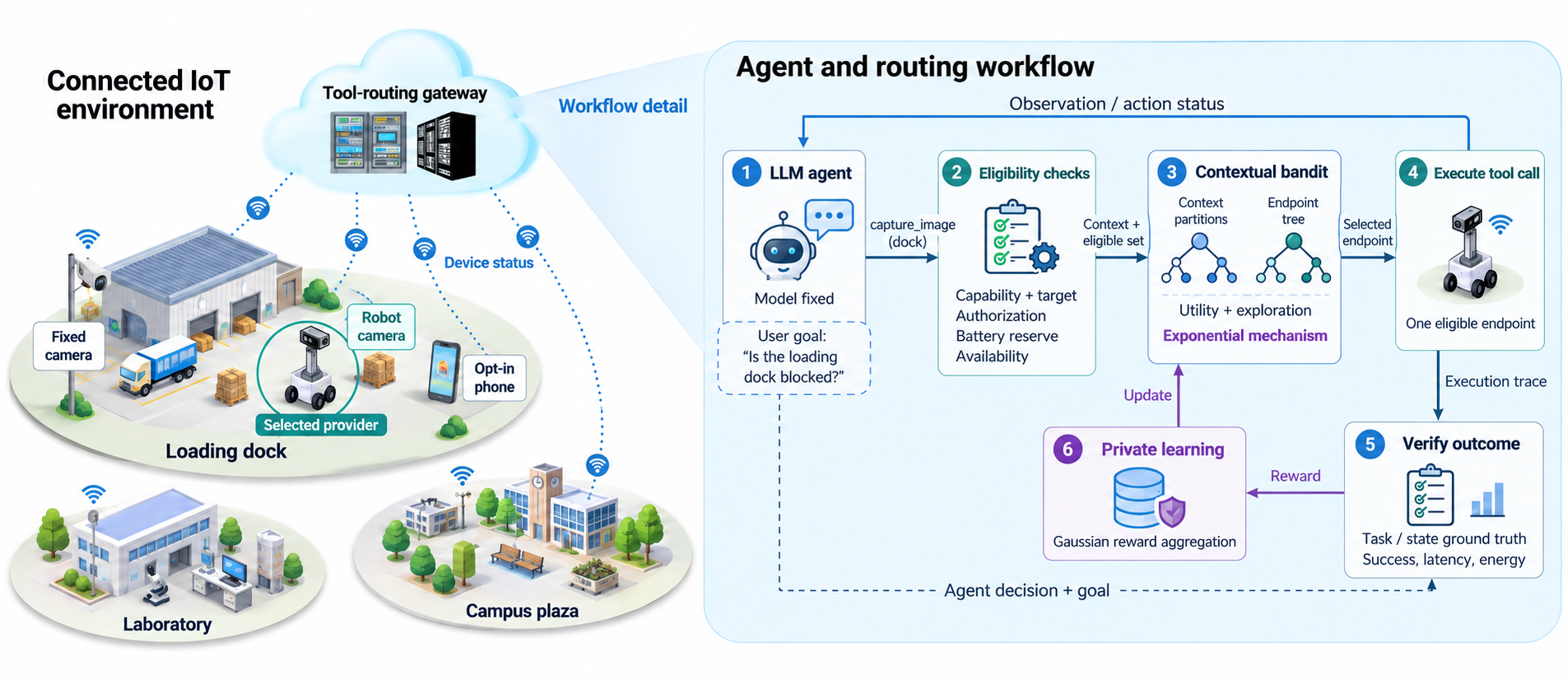}
    \caption{Tool-grounding workflow. Steps 1--6 form the routing loop for one tool call: the router selects an endpoint, the call is executed and verified, and the reward updates the learner. The top arrow returns the observation to the LLM agent. The dashed line marks the agent's own check of its decision against the user goal; this check stays with the LLM and is not part of the evaluated loop. The evaluation covers provider selection in simulation rather than the complete agent--device loop.}
    \label{fig:system_diagram_times}
\end{figure*}

Another line of work improves these decisions from outcomes, by adapting retrieval during deployment or offline~\cite{pan2025online,chen2026oats}, or by optimizing agent workflows, exploration, skill reuse, and self-evolving training loops~\cite{nie2026understanding,ye2026look,yao2026skillrise,wang2026agentopsd,chen2026falsefrontiers}. These methods adapt the agent or its retriever; our learner keeps the LLM fixed and adapts only the choice of physical endpoint. Runtime security studies are complementary to per-call eligibility~\cite{jiang2026attacksurface,jiang2026chaincaps}.

\subsection{Physical Grounding and IoT Scheduling}
SayCan, IoT-Brain, and intent-aware frameworks add skill affordances, spatial relationships, and capability checks to physical plans~\cite{ichter2023saycan,zhou2026iotbrain,jin2026dsia}, and adaptive environmental monitoring selects sensors under energy constraints~\cite{liu2026environmental}. These approaches determine which plans are executable; we instead learn how useful each eligible provider is for a given request. Following studies that show aggregate scores can hide regime-specific failures~\cite{wang2026timeseriesfoundationmodel,sun2026accuracymeasuringbiasacknowledgment}, we state the scope of our evaluation explicitly in Section~\ref{sec:results}.

\subsection{Contextual Bandits for Provider Selection}
Contextual bandits have been used to learn IoT service utility and crowdsourcing participation~\cite{wanigasekara2016bandit,klos2018crowdsourcing}, to scale to large candidate sets through hierarchical and similarity-based partitions~\cite{bubeck2011xarmed,slivkins2014similarity}, and to route prompts among LLMs~\cite{poon2026multillm}. We carry this local learning to physical endpoints, where eligibility is a hard constraint rather than a feature.

\subsection{Private Online Learning for IoT}
Privacy in IoT learning has been studied mainly for federated model training~\cite{hou2026fedcsa,hou2026asyncfedsar,hou2026privatesar,venkatesh2026dpprox}. These methods protect model updates; our learner protects the reward statistics that drive routing. Private tree-based IoT service selection is the closest methodological precedent~\cite{zhao2019privatetree}, and private bandit studies show that guarantees depend on which records are protected and what an observer can see~\cite{shariff2018private,zheng2020local}. Our aggregation follows these principles and claims no privacy for raw observations or for the complete adaptive transcript.

\section{Proposed Method}
\label{sec:method}

Tool grounding proceeds in five steps: check device eligibility, locate the relevant partitions, select a provider, verify execution, and update the learner. Fig.~\ref{fig:system_diagram_times} shows the system workflow.

\subsection{Receiving the Call and Checking Eligibility}

We consider a connected deployment whose registry holds the endpoints $\mathcal D_n$ available at round $n=1,\ldots,N$, where $N$ is the horizon and each endpoint $p\in\mathcal D_n$ is a device--capability pair. The agent supplies the requested capability, the physical target, and a context $e_n\in\mathcal E=[0,1]^{\kappa_e}$ whose $\kappa_e$ features describe target location, task intent, required quality, service duration, and deadline. Each endpoint carries current features $z_n(p)\in\mathcal P=[0,1]^{\kappa_p}$ describing location, quality, latency, battery state, availability, and operating constraints. Both feature maps are fixed before learning, and capability and authorization are checked explicitly rather than expressed as numerical similarity.

The router handles calls sequentially and first removes devices incompatible with the requested operation and target. Let $g_n(p)=1$ indicate capability, target, permission, and network compatibility. The eligible set is
\begin{equation}
\begin{split}
\mathcal D_n^{\rm elig}=\bigl\{p\in\mathcal D_n:\;&g_n(p)=1,\\
&\widehat\tau_n(p)\leq\min\{\Delta_n,W_n(p)\},\\
&q_n(p)-\widehat E_n(p)/\Gamma_p\\
&\qquad\geq V_p^{\rm min}+U_n(p)\bigr\}.
\end{split}
\label{eq:eligible}
\end{equation}
Here $\widehat\tau_n(p)$ and $\widehat E_n(p)$ are the predicted completion time and call energy of endpoint $p$, $\Delta_n$ is the deadline allowance carried by the request, and $W_n(p)$ is the time for which $p$ remains available from the current round. For battery-powered devices, $q_n(p)\in[0,1]$ is the state of charge, $\Gamma_p$ the battery capacity in the same energy units as $\widehat E_n(p)$, $V_p^{\rm min}$ the reserve floor the device keeps in hand, and $U_n(p)$ the charge fraction already committed to scheduled operations. Mains-powered devices omit the last check. Because these tests rely on predictions, the selected endpoint rechecks permission and reserves immediately before execution and locally enforces its admitted energy allowance. Rejection, interruption, or timeout counts as failure, and if no endpoint is eligible the router reports unavailability.

\subsection{Locating the Context and Endpoint Partitions}

\begin{figure}[t]
\centering
\includegraphics[width=\columnwidth,trim=0 0 0 4.875bp,clip]{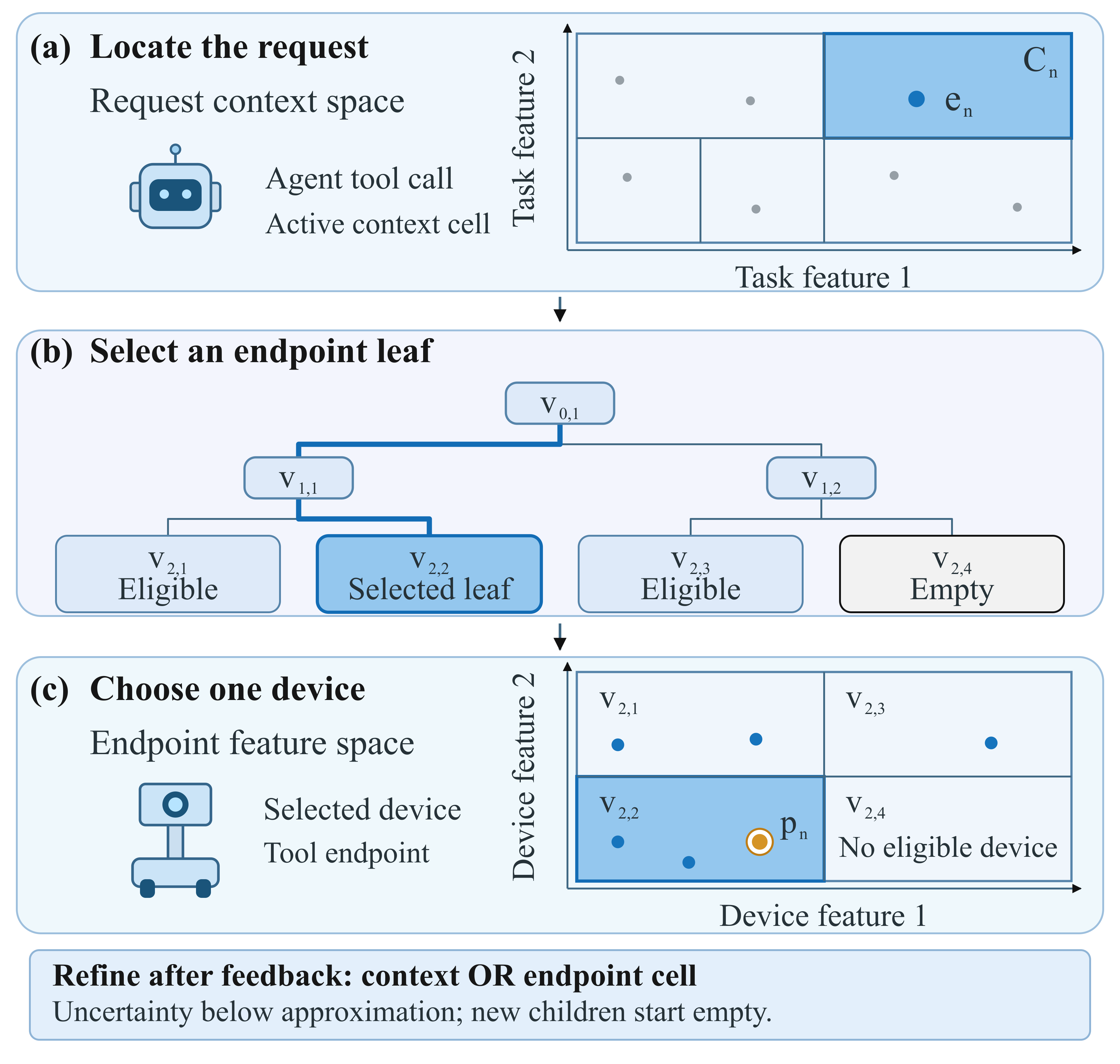}
\caption{Context-specific endpoint partitioning. (a) The current request enters context cell $C_n$. (b) Its endpoint hierarchy organizes eligible providers; the highlighted route ends at the sampled leaf. (c) That leaf is a region of endpoint feature space, from which one eligible device is selected. The coordinate plots show illustrative two-dimensional slices. Refinement splits one context or endpoint cell at a time; new children start without feedback.}
\label{fig:partitions}
\end{figure}

Within each capability class, a binary tree $\Lambda^E$ partitions the request space, and each active context cell $C$ carries its own endpoint tree $\Lambda_C^P$. This two-sided structure~\cite{liu2022private} shares experience among similar requests while keeping provider estimates specific to the conditions under which they were observed, so a camera that is reliable for obstruction detection is not credited with the same utility for reading a label. Fig.~\ref{fig:partitions} shows how a request is mapped to a context cell and then to that cell's endpoint hierarchy.

The router locates $e_n$ in cell $C_n$ and places the eligible endpoints in that cell's endpoint tree using their current features, while historical feedback retains the features recorded at selection. Endpoints may therefore move between leaves as their state changes. Only leaves containing eligible endpoints enter the candidate set $\mathcal V_n$. Each endpoint leaf $v=(l,i)$, indexed by its depth $l$ in the tree and its position $i$ at that depth, has two children that partition its region: cyclic axis bisection fixes where a split occurs, learning determines when the children are activated, and depth caps limit memory.

Let $M_e$ and $M_p$ be fixed weighted maximum-norm metrics on $\mathcal E$ and $\mathcal P$, and let $D_C$ and $D_v$ be the diameters of a context cell $C$ and an endpoint leaf $v$ under those metrics. Within each compatible capability class, we assume that changing only the request context changes the mean reward by at most $L_e$ times the context distance, and that changing only the endpoint features gives the analogous bound with $L_p$ and the endpoint distance, where $L_e$ and $L_p$ are fixed nonnegative smoothness constants~\cite{bubeck2011xarmed,slivkins2014similarity}.

Applying these bounds successively limits the mean-reward difference between any two interactions in the same cell pair to $L_eD_C+L_pD_v$. This allowance enters the index in~\eqref{eq:grounding-index}, and its two components also guide refinement in~\eqref{eq:refinement-rule}. The assumptions must be assessed for the chosen features and the fixed LLM, since shared capability or physical proximity alone does not establish them. For example, two nearby cameras can differ sharply in usefulness when a wall blocks only one of them, so $L_e$ and $L_p$ are modeling choices rather than verified properties.

\vspace{-5pt}
\subsection{Selecting a Physical Provider}

Each cell pair $(C,v)$ stores a completed-feedback count $\Omega_{C,v}$, the number of finished calls attributed to it, and a private mean $\widetilde\mu_{C,v}$ supplied by the noisy aggregation described below. For visited pairs, the router computes the sampling width $w_{C,v}$, the noise width $\nu_{C,v}$, and the selection index $Q_{C,v}$ as
\begin{equation}
\begin{aligned}
w_{C,v}&=\sqrt{2\ln(4N^2/\beta)/\Omega_{C,v}},\\
\nu_{C,v}&=\sigma\sqrt{2J\ln(4N^2/\beta)}/\Omega_{C,v},\\
Q_{C,v}&=\widetilde\mu_{C,v}+w_{C,v}+\nu_{C,v}+L_eD_C+L_pD_v.
\end{aligned}
\label{eq:grounding-index}
\end{equation}
Here $\beta\in(0,1)$ is a failure probability, while the noise scale $\sigma$ and the number of temporal-tree levels $J$ are defined below. The two widths cover the two sources of error in $\widetilde\mu_{C,v}$: $w_{C,v}$ accounts for finite sampling and $\nu_{C,v}$ for the injected noise. A union bound over at most $N$ stream births and $N$ prefix lengths controls both errors with probability at least $1-\beta$.

When eligible leaves still lack feedback, the router samples uniformly among those unvisited leaves. Otherwise, it samples a leaf $v_n$ using
\begin{equation}
\Pr(v_n=v)=\frac{\exp(Q_{C_n,v}/\theta_n)}
{\sum_{u\in\mathcal V_n}\exp(Q_{C_n,u}/\theta_n)}.
\label{eq:randomized-routing}
\end{equation}
The temperature $\theta_n>0$, for example $\theta_0/\sqrt n$, controls how sharply the distribution concentrates on larger indices. Sampling operates on sanitized scores, but the temperature is not a calibrated privacy budget and the rule adds no separate DP guarantee. The router then samples one eligible endpoint $p_n$ uniformly within the chosen leaf and invokes the requested operation.

\subsection{Executing the Call and Verifying Its Outcome}

In the full workflow, the observation is returned to the fixed LLM and an independent verifier supplies a binary success label $Y_n\in\{0,1\}$. Deployment labels could come from annotated scenes or actuator acknowledgments. The simulation instead samples success indicators directly, without a verifier, because neither a valid payload nor an LLM self-assessment constitutes ground truth.

The algorithm waits for final feedback before updating. Verifier error also propagates into what is learned: with false-positive rate $a$ and false-negative rate $b$, a true success probability $\pi$ is learned as $a+(1-a-b)\pi$, which can change provider rankings.

Let $\tau_n$ and $E_n$ denote the measured latency and energy of the completed call, including communication, and let $\bar\tau_n=\min\{\tau_n/\Delta_n,1\}$ and $\bar E_n=\min\{E_n/E_{\rm ref},1\}$ normalize them, where $E_{\rm ref}>0$ is a capability-specific energy scale. The reward is
\begin{equation}
r_n=Y_n\exp(-\lambda_\tau\bar\tau_n-\lambda_E\bar E_n)\in[0,1],
\label{eq:grounding-reward}
\end{equation}
where the latency and energy penalty weights $\lambda_\tau,\lambda_E\geq0$. A failed call receives zero, and a successful call earns more when it finishes faster and consumes less energy. Only the selected endpoint's outcome is observed, and a retry starts a new round.

Within a stationary regime, assume $\mathbb E[r_n\mid\mathcal F_n,p_n=p]=\mu(e_n,z_n(p))$, where $\mu$ is the conditional mean reward of a request--endpoint pair and $\mathcal F_n$ is the history and current information available before selection. Battery depletion enters through the features $z_n(p)$ and the eligible set, while the mapping $\mu$ is assumed fixed within a regime; a change that alters $\mu$ itself, such as a new occlusion, starts a new regime. The objective is to minimize the regret
\begin{equation}
\begin{aligned}
R_N=\mathbb E\biggl[\sum_{n\in\mathcal I_N}\biggl(
&\max_{p\in\mathcal D_n^{\rm elig}}\mu(e_n,z_n(p))\\
&-\mu(e_n,z_n(p_n))\biggr)\biggr],
\end{aligned}
\label{eq:grounding-regret}
\end{equation}
where $\mathcal I_N$ collects the rounds in which at least one endpoint is eligible. The comparator uses the learner's realized device states at each round rather than optimizing a separate future battery trajectory, and rounds without an eligible endpoint are evaluated separately.

\subsection{Updating Private Statistics and Refining the Trees}

A trusted feedback service receives the verified reward and appends it once to the selected cell pair's stream. A temporal binary tree stores the reward sums over intervals, each completed node $j$ receives one independent perturbation $Z_j\sim\mathcal N(0,\sigma^2)$ that is reused in subsequent queries, and any prefix combines at most $J=1+\lceil\log_2N\rceil$ nodes. We write $\widetilde S_{C,v}$ for the resulting noisy prefix sum of a cell pair's stream. This aggregation tree is separate from the feature-partition trees.

\noindent\textit{Proposition 1 (fixed-assignment reward privacy).}
Fix a reward-independent schedule containing metadata, stream assignments, local record order, and release times. Neighboring histories differ in one reward in $[0,1]$ and share that schedule. For privacy parameters $\varepsilon>0$ and $0<\delta<1$, let $\sigma$ be at least the noise scale at which the Gaussian mechanism with $\ell_2$-sensitivity $\sqrt J$ is $(\varepsilon,\delta)$-DP under analytic calibration~\cite{balle2018improving}. Then the joint release of all noisy prefixes and fixed counts is $(\varepsilon,\delta)$-DP. For $\varepsilon<1$, the closed form $\sigma\geq\sqrt{2J\ln(1.25/\delta)}/\varepsilon$ is sufficient~\cite{dwork2014privacy}.

\noindent\textit{Proof.} Each reward belongs to one stream and, at each of the $J$ temporal levels, to one interval sum. Replacing a single reward therefore changes at most $J$ coordinates of the vector of all node sums, taken across all fixed streams, each by at most one, so this vector has $\ell_2$-sensitivity at most $\sqrt J$. Adding independent $\mathcal N(0,\sigma^2)$ noise to every node is a single Gaussian mechanism with this sensitivity, which gives the result~\cite{balle2018improving}. Reused prefixes and the clipped mean
\begin{equation}
\widetilde\mu_{C,v}=\operatorname{clip}_{[0,1]}
\bigl(\widetilde S_{C,v}/\Omega_{C,v}\bigr)
\label{eq:private-mean}
\end{equation}
are post-processing. Fixing the schedule independently of rewards is essential, since conditioning afterward on a reward-dependent trace is a different experiment. \hfill$\blacksquare$

The adversary in this model is an honest-but-curious statistical observer, such as an operator dashboard or a shared tuning pipeline, that receives the released aggregates and tries to infer individual reward values. The trusted feedback service sees raw labels, costs, rewards, and exact sums, whereas the statistical observer sees only fixed metadata, counts, and sanitized aggregates. The proposition does not cover the complete live transcript, because rewards can change later routing, device states, assignments, and refinement, which may disclose information through subsequent behavior. Raw requests, payloads, and verifier records lie outside this guarantee.

After updating the statistics, the router tests
\begin{equation}
w_{C,v}+\nu_{C,v}\leq\max\{L_eD_C,L_pD_v\}.
\label{eq:refinement-rule}
\end{equation}
Once uncertainty falls below the larger approximation term, the router splits that side: the context cell if $L_eD_C>L_pD_v$, and the endpoint cell otherwise. Context children receive fresh endpoint roots, new children start empty with no replay of raw feedback, and unchanged cells retain their statistics. At most one split occurs per round, and a blocking depth cap postpones refinement, leaving an approximation floor. Changes to the LLM, verifier, or tool semantics start a new regime.

\subsection{Finite-Horizon and Resource Bounds}
\label{subsec:bounds}

Let $H=\ln(4N^2/\beta)$, let $S_N$ count all nonempty feedback streams, including retired ones, let $K_n=|\mathcal V_n|$ be the number of eligible leaves at round $n$, and let $b_n=L_eD_{C_n}+L_pD_{v_n}$ be the cell bias of the selected pair. Under the confidence assumptions above, the stated algorithm satisfies
\begin{equation}
\begin{split}
R_N\leq\beta N+\mathbb E\bigl[&S_N+4\sqrt{2HS_NN}\\
 &+2\sigma\sqrt{2JH}\,S_N(1+\ln N)\\
 &+2\!\sum_{n\in\mathcal I_N}b_n
 +\!\sum_{n\in\mathcal I_N}\theta_n\ln K_n\bigr].
\end{split}
\label{eq:finite-bound}
\end{equation}
Each term matches one part of the argument: a first visit to a stream costs at most one; on later visits optimism bounds the regret by twice the sampling width, the noise width, and the cell bias, plus the deficit from choosing below the largest index, which the entropy bound limits to $\theta_n\ln K_n$. Summing the inverse-square-root and inverse-count widths gives~\eqref{eq:finite-bound}, and the bad event costs at most $\beta N$.

The bound identifies the quantities that drive the loss, and shows that the refinement rule is what controls the cell-bias term, but it can be vacuous when coarse cells leave a linear bias or when resets increase $S_N$, and it establishes neither convergence nor superiority to UCB1 or LinUCB. With $C_0$ capability roots, at most one split per round allocates $O(C_0+N)$ spatial nodes, and the prefix stacks require $O(S_NJ)$ numerical storage, excluding payloads and logs. Routing one call takes $O(h_e+mh_p+K_nJ)$ tree and aggregate operations for $m=|\mathcal D_n|$ endpoints and tree depths $h_e$ and $h_p$, in addition to feature and eligibility computation. These are complexity bounds, not measured runtime.

\section{Numerical Results}
\label{sec:results}

\begin{figure*}[!t]
    \centering
    \includegraphics[width=\textwidth]{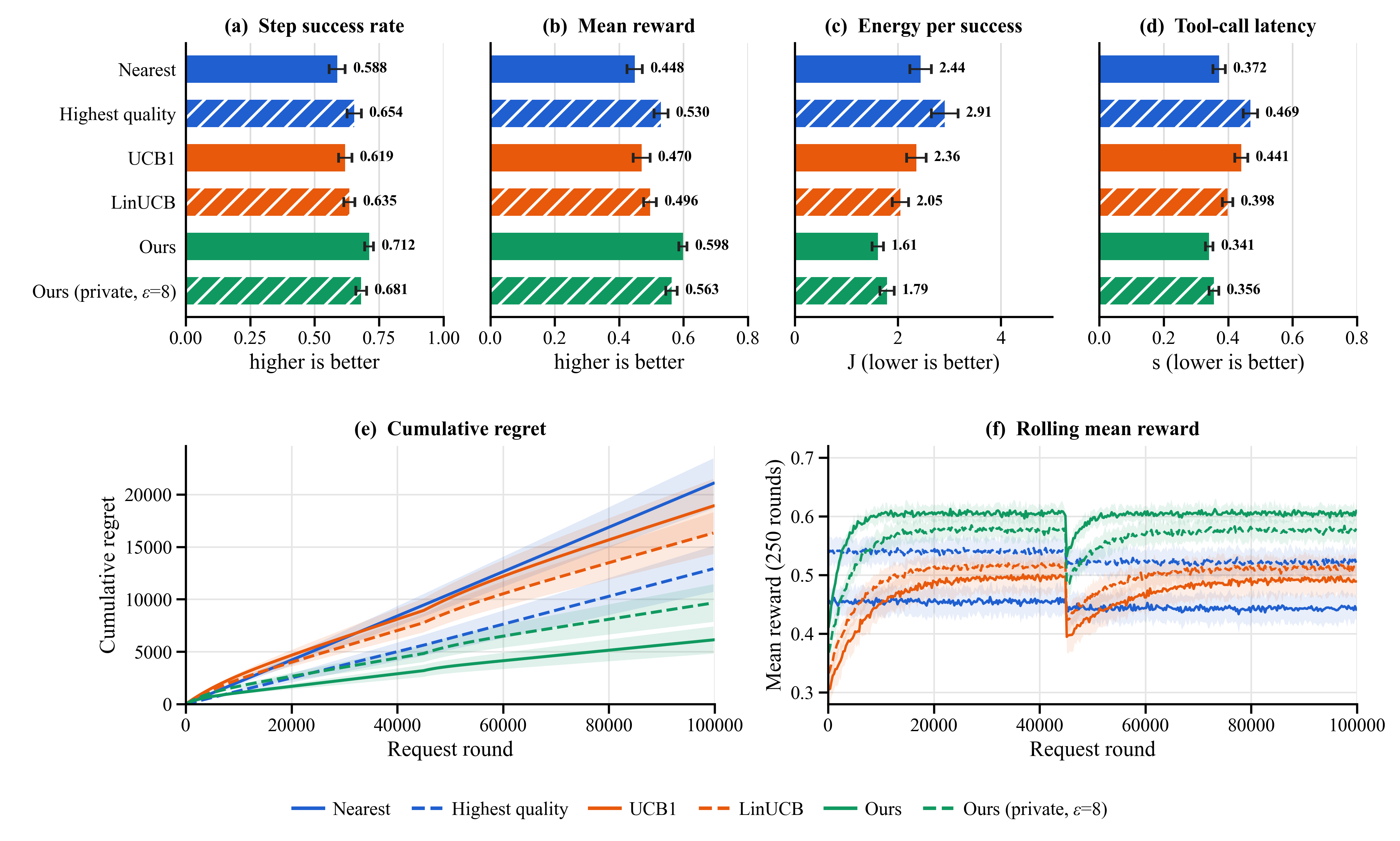}
    \caption{Performance over $N=100{,}000$ requests: (a) step success rate,
    (b) mean reward, (c) device energy per successful step, (d) tool-call latency,
    (e) cumulative regret, and (f) rolling mean reward over a 250-request window.
    Bars and lines are five-seed means; error bars and shaded bands show $\pm1$
    sample standard deviation. The noisy variant uses $\varepsilon=8$. Energy and
    latency are simulated device-side quantities and exclude LLM inference and
    router computation.}
    \label{fig:results}
\end{figure*}

\subsection{Simulation Scenario and Basic Settings}

The simulated workload asks whether a loading dock or passage is blocked, and each \texttt{capture\_image(area)} request is routed to one eligible camera. Responses and success labels are generated by the simulator, without image processing, LLM inference, or physical verification, so the experiment evaluates provider selection rather than complete agent tasks.

Each of the five environment seeds contains $N=100{,}000$ requests, and all percentage comparisons are computed from the reported means. No external dataset is used: requests, device states, and outcomes are all generated by the simulator. Table~\ref{tab:settings} lists the main settings.

The campus has 6 zones with 5 targets each and 300 endpoints, and a call has 52--72 eligible endpoints on average. The success probability of a call is a logistic function of quality match, distance to the target, a latent line-of-sight visibility, and a night-time penalty for cameras. Visibility is set by 16 thin walls between devices and targets. Visibility and lighting are hidden from every router, so advertised quality and distance only partly predict success. Latency and energy are drawn per device class, and battery devices discharge on each call and recharge at a fixed rate.

In the implementation, the widths in~\eqref{eq:grounding-index} use the scaled forms $w_{C,v}=c_w\sqrt{\ln(1+n)/\Omega_{C,v}}$ and $\nu_{C,v}=c_w\sqrt J\,\sigma/\Omega_{C,v}$, because the factor $\ln(4N^2/\beta)\approx27$ would otherwise dominate the index. This changes exploration only, not the released statistics or their privacy, and the bound in~\eqref{eq:finite-bound} applies to the stated widths.

\begin{table}[t]
\caption{Simulation and Learner Settings}
\label{tab:settings}
\centering
\footnotesize
\setlength{\tabcolsep}{3pt}
\renewcommand{\arraystretch}{1.08}
\begin{tabular}{p{0.44\columnwidth}p{0.50\columnwidth}}
\hline\hline
\textbf{Parameter} & \textbf{Value} \\
\hline
Zones; targets per zone; endpoints & 6; 5; 300 \\
Eligible endpoints per call & 52--72 (by seed) \\
Advertised quality & mean 0.58--0.85, s.d. 0.09--0.18 \\
Latency; energy per call & 0.20--1.10~s; 0.4--5.0~J \\
Battery capacity $\Gamma_p$; reserve $V_p^{\rm min}$ & $U(4000,12000)$~J; $U(0.20,0.45)$ \\
Committed usage $U_n(p)$; recharge & $U(0.02,0.12)$; $1.2\times10^{-5}$ per round \\
Walls; night penalty (cameras) & 16 thin walls; $U(0.05,2.2)$, 20:00--06:00 \\
Success logit weights & bias $-0.55$, quality 1.9, distance $-2.6$, visibility 2.2, lighting $-2.4$ \\
$\Delta_n$; $\lambda_\tau$; $\lambda_E$; $E_{\rm ref}$ & $U(0.8,3.0)$~s; 0.6; 0.4; 6.0~J \\
$\kappa_e$; $\kappa_p$ & 6; 6 \\
$L_e$; $L_p$; $c_w$; $\beta$ & 0.20; 0.20; 0.15; 0.05 \\
Split threshold (endpoint; context) & 40; 15{,}000 calls \\
Initial / maximum depth (context; endpoint) & 2 / 12; 16 / 16 \\
Temperature $\theta_n$ & $\max(0.10/\sqrt n,\,10^{-3})$ \\
$J$; $\delta$; $\sigma$ at $\varepsilon=8$ & 18; $10^{-5}$; 2.57 \\
UCB1 constant; LinUCB $\alpha$, ridge & 0.6; 0.35, 1.0 \\
\hline\hline
\end{tabular}
\end{table}

\subsection{Benchmark Design}

\subsubsection{Baseline Algorithms}

All six methods select among eligible endpoints and differ only in how a provider is chosen.

\noindent\textbf{Nearest eligible device (Nearest).} Selects the endpoint closest to the requested area, using location alone.

\noindent\textbf{Highest advertised quality (Highest quality).} Selects the endpoint with the highest advertised observation quality, testing metadata without learning.

\noindent\textbf{UCB1~\cite{auer2002finite}.} Selects the largest empirical mean reward plus an exploration allowance, pooling feedback across requests without context.

\noindent\textbf{LinUCB~\cite{li2010contextual}.} Selects by predicted reward plus an uncertainty allowance under a model linear in the request--endpoint features, testing contextual learning without local partitions.

\noindent\textbf{Proposed method without noise (Ours).} Uses the proposed partitions with unperturbed aggregates, as the ablation reference for the cost of noise.

\noindent\textbf{Proposed method with aggregation noise (Ours private).} Uses noisy aggregates and their uncertainty allowance in selection and refinement, at $\varepsilon=8$ and $\delta=10^{-5}$, which give $J=18$ and a per-node noise scale $\sigma=2.57$ that meets the calibration of Proposition~1. It is labelled Ours (private, $\varepsilon=8$) in Fig.~\ref{fig:results}. In these runs, stream assignments and refinement depend on earlier rewards, so the runs do not meet the fixed-schedule condition of Proposition~1. The gap between the two variants therefore measures the utility cost of this noise level, not a privacy guarantee for the run.

\subsubsection{Performance Measures}

\noindent\textbf{Simulated step success rate.} The fraction of requests receiving a successful simulated observation.

\noindent\textbf{Mean reward.} Equation~\eqref{eq:grounding-reward} discounts success by latency and energy and assigns failure zero, so a higher value means better cost-adjusted service.

\noindent\textbf{Cumulative regret.} Equation~\eqref{eq:grounding-regret}, accumulated against the best eligible endpoint at the learner's realized device state.

\noindent\textbf{Rolling mean reward.} A 250-request average exposing exploration and recovery.

\noindent\textbf{Device energy per successful step.} Joules of device energy per successful service; under total-attempt accounting the numerator includes failed calls.

\noindent\textbf{Tool-call latency.} Simulated device-side delay in seconds. Both cost measures exclude LLM inference, router computation, and verification overhead.

\subsection{Comparison with Baseline Algorithms}

\subsubsection{Success of the Selected Provider}

The two proposed variants achieve the highest mean step success rates among the evaluated methods, as shown in Fig.~\ref{fig:results}(a). Ours reaches $71.2\%$ and Ours private $68.1\%$, so even the noisy variant exceeds UCB1 at $61.9\%$ and LinUCB at $63.5\%$ by $6.2$ and $4.6$ percentage points. Highest quality is the strongest baseline at $65.4\%$, and Ours private improves on it by a further $2.7$ percentage points. This margin is small, and five seeds are not enough to test its significance.

Two mechanisms plausibly account for this ranking: context partitions separate the request-dependent utility that UCB1 pools, while local estimates capture relationships that are nonlinear in LinUCB's features. The comparison does not isolate features, tuning, or the two partitions, so these remain explanations rather than attributions.

\subsubsection{Reward and the Success--Cost Balance}

The same ordering holds once service cost is taken into account. Fig.~\ref{fig:results}(b) reports mean rewards of $0.598$ for Ours and $0.563$ for Ours private, the latter exceeding UCB1's $0.470$ by approximately $19.8\%$ and LinUCB's $0.496$ by $13.5\%$, and also exceeding Highest quality, the strongest baseline on this measure.

Because success and reward rise together, the additional successes are not obtained by spending more time or energy, as expected from the combined objective in~\eqref{eq:grounding-reward}.

\subsubsection{Device Energy Efficiency}

Fig.~\ref{fig:results}(c) shows energy per successful step of $1.61$~J for Ours and $1.79$~J for Ours private. LinUCB is the most efficient baseline at $2.05$~J, so the noisy variant reduces the reported metric by $12.7\%$; against Highest quality at $2.91$~J the reduction is $38.5\%$.

Highest quality combines a strong success rate with the highest energy cost, which is why advertised quality alone is an incomplete objective; the energy penalty instead favors useful providers with lower service cost. Under total-attempt accounting fewer failures also lower the ratio, and the averages do not separate this from lower per-call energy.

\subsubsection{Tool-Call Latency}

Fig.~\ref{fig:results}(d) reports latencies of $0.341$~s for Ours and $0.356$~s for Ours private, placing the noisy variant $10.6\%$ below LinUCB's $0.398$~s and $4.3\%$ below Nearest, the fastest baseline at $0.372$~s.

Proximity need not minimize response time, since device processing and communication also contribute. The noisy variant is slightly slower than Ours, likely because aggregation noise keeps it trying less-tested endpoints, some of which respond more slowly. The measure says nothing about deadline misses or tail latency, and it excludes routing computation, whose cost is bounded in Section~\ref{subsec:bounds}.

\subsection{Learning Behavior over the Request Sequence}

\subsubsection{Cumulative Utility Loss}

The advantage persists across the request sequence. At the final round in Fig.~\ref{fig:results}(e), cumulative regret is approximately $6.1\times10^3$ for Ours and $9.7\times10^3$ for Ours private, against $1.63\times10^4$ for LinUCB, so the noisy variant loses roughly $40\%$ less utility overall while still paying for early exploration against the feasible per-call oracle. These finite-horizon curves do not by themselves establish vanishing average regret or convergence to that oracle.

\subsubsection{Recent Reward and Recovery}

Fig.~\ref{fig:results}(f) shows both proposed variants improving during the early rounds, with Ours settling near $0.60$ and Ours private near $0.57$--$0.58$. The gap between them is the continuing utility cost of noisy aggregation. After the transient near request $45{,}000$, both return to higher rolling rewards than the baselines, and Ours recovers earlier.

\subsection{Privacy--Utility Tradeoff and Discussion}

Collecting the two comparisons above, aggregation noise costs $3.1$ percentage points of step success and $5.9\%$ of mean reward, the expected shape of a privacy--utility tradeoff. The price is modest here, since the noisy variant still has better means than all four baselines on every reported measure. Taken together, the results suggest that learning request-dependent provider utility is useful for this simulated loading-dock workload, at a utility cost when reward aggregates are protected.

\subsection{Limitations}

This study has several limitations. First, the evaluation uses our own simulator, in which success labels are sampled directly. It does not include real devices, images, LLM inference, or a verifier, so the results describe router behavior in this workload rather than gains in deployment. Second, the comparison covers four baselines over five seeds. It does not include contextual zooming~\cite{slivkins2014similarity}, Thompson sampling, or private tree-based selection~\cite{zhao2019privatetree}, and it does not separate the effects of the context partitions, the endpoint trees, and the feature design. Third, the regret bound relies on the smoothness and stationarity assumptions of Section~\ref{sec:method}, can be vacuous when cells stay coarse, and was not tested for sensitivity to $L_e$ and $L_p$. Under the simulated recharge rate, batteries stayed close to full charge, so the reserve check in~\eqref{eq:eligible} did not remove endpoints in practice and the energy side of eligibility was not exercised. Fourth, Proposition~1 covers released reward statistics under a fixed, reward-independent schedule. It does not cover the adaptive router used in our runs, which we evaluated only at $\varepsilon=8$, and we report no sweep over $\varepsilon$ or empirical leakage test.

\section{Conclusion}
\label{sec:conclusion}

In this paper, we presented an online tool-grounding framework for selecting physical device providers for LLM agents in connected IoT systems. Adaptive context and endpoint partitions reuse execution feedback to learn request-dependent device utility while accounting for authorization, battery reserves, availability, and deadlines. The analysis characterizes finite-horizon regret under the stated assumptions. Gaussian reward aggregation provides differential privacy for released statistics under fixed, reward-independent assignments. In the simulated loading-dock workload, both proposed variants achieve higher mean step success and mean reward and lower cumulative regret than the four evaluated baselines, with lower reported device energy per successful step and tool-call latency. Adding noise reduces success and reward, showing the performance cost of reward-value protection. Future work will calibrate the success model with a vision-language model on real images, test the router on trace-driven and physical device data, add matched component studies and stronger baselines, and extend the privacy accounting to the full adaptive transcript. Delayed feedback and end-to-end agent execution under changing operating conditions also remain open.

\end{document}